\documentclass[final]{hooml2022} 

\usepackage[utf8]{inputenc} 
\usepackage[T1]{fontenc}    
\usepackage{hyperref}       
\usepackage{url}            
\usepackage{booktabs}       
\usepackage{amsfonts}       
\usepackage{nicefrac}       
\usepackage{microtype}      
\usepackage{float}
\usepackage{subcaption}

\DeclareMathOperator*{\argmin}{argmin}

\vspace{-15mm}
\title{Improving Levenberg-Marquardt Algorithm for Neural Networks}

\vspace{-15mm}

\author{\Name{Omead Pooladzandi} \Email{opooladz@ucla.edu}\\
\Name{Yiming Zhou} \Email{yimingz0416@ucla.edu}\\
\addr{UCLA 420 Westwood Plaza Los Angeles, CA 90095}}

\begin{document}

\maketitle
\vspace{-5mm}
\begin{abstract}
We explore the usage of the Levenberg-Marquardt (LM) algorithm for regression (non-linear least squares) and classification (generalized Gauss-Newton methods) tasks in neural networks. We compare the performance of the LM method with other popular first-order algorithms such as SGD and Adam\citep{kingma2014adam}, as well as other second-order algorithms such as L-BFGS \citep{Liu89onthe}, Hessian-Free \citep{martens2010deep} and KFAC\citep{martens2015optimizing}. We further speed up the LM method by using adaptive momentum, learning rate line search, and uphill step acceptance.

\end{abstract}

\section{Introduction}

First-order methods are easy to use and require moderate computational cost per iteration for optimizing neural networks. However, their hyper-parameters are cumbersome to tune, and the convergence rate can decrease as the neural network (NN) size becomes large. Second-order methods leverage the Hessian matrix (i.e. the curvature of the objective function) and enjoy faster convergence, robustness, and affine invariance. In a single step, instead of just advancing towards the minimum, it steps towards the global minimum under the assumption that the function is quadratic and its second-order expansion is a good approximation. This allows the optimizer to make big steps in low-curvature scenarios and small steps in high-curvature scenarios. However, computing and storing the Hessian $H = \nabla ^2f(x)$ and its inverse is expensive in computation (time complexity $O(n^3)$) and memory (space complexity $O(n^2)$). When the number of parameters in a NN becomes large it is prohibitively expensive to compute the full Hessian inverse. Several approaches, such as the Gauss-Newton method, the Limited-Memory Broyden–Fletcher–Goldfarb–Shanno (LBFGS) method, the Hessian-Free method, and the KFAC method \citep{Liu89onthe, martens2010deep,martens2015optimizing}, have been proposed to approximate curvature information making second-order algorithms more efficient. 

In this work, we provide a version of the Levenberg-Marquardt (LM) Algorithm for Neural Networks which utilizes adaptive momentum, learning rate line search, and uphill step acceptance to increase convergence as well as performance. We compare the modified LM method to LBFGS, HF, KFAC to SGD and Adam for training a CNN on the MNIST dataset and an MLP on a noisy Sine regression task.  For more on each method see Appendix \ref{appendix}. 

\section{LM Algorithm for Neural Networks}

There have been many updates to the standard Levenberg-Marquardt algorithm. In this section, we consider some of these extensions and combine the methods to get an updated version of LM.   
\subsection{Gauss-Newton Method}

Gauss-Newton (GN) method was originally proposed to solve a nonlinear least square problem, where $r: \mathbb{R}^n \rightarrow \mathbb{R}^m$ denotes the residual $r(x) = [r_1(x), ..., r_m(x)]^T$ of a $n$-vector $x$.
\vspace{-3mm}
\begin{equation}
    \min \quad g(x)= \Vert r(x) \Vert_2^2 = \sum_{i=0}^{m}r_i(x)^2
    \vspace{-3mm}
\end{equation}
To optimize the function above, we can take a Newton step, which will do the following update
\vspace{-3mm}
\begin{equation}
    v_{\text{nt}} = -\nabla^2g(x)^{-1}\nabla g(x) = -(J^T J + \sum_{i=1}^{m}r_i(x)\nabla^2r_i(x))^{-1}J^Tr
    \vspace{-3mm}
\end{equation}
where $J = \frac{\partial r(x)}{\partial x}$ is the Jacobian matrix of $r(x)$. However in the Gauss-Newton step, we discard the second term in the inverse and simply do
\vspace{-3mm}
\begin{equation}
    v_{\text{gn}} = -(J^TJ)^{-1}J^Tr.
    \vspace{-3mm}
\end{equation}
The Gauss-Newton step doesn't require second derivatives and always provides a descent direction when $J$ has full column rank; because $H=2J^TJ$ is positive definite.

The Levenberg-Marquardt algorithm is a variant of the Gauss-Newton method which addresses the indefiniteness of the approximated Hessian by adding a regularization term $\lambda$
\vspace{-3mm}
\begin{equation}
    v_{\text{gn}} = -(J^TJ + \lambda I)^{-1}J^Tr.
    \vspace{-3mm}
\end{equation}

Here $\lambda$ defines a trust region where the quadratic approximation is a good fit. It blends linear and quadratic approximations. When $\lambda = 0$, the update rule is completely based on the second-order approximation, when $\lambda$ is very large, the rule is completely based on the first-order approximation.

\subsection{Generalized Gauss-Newton Method}

Consider a neural network (NN) with $L+1$ layers, at the $l$-th layer, given the outputs from the preceding layer $x^{(l-1)}$ as input, $x^{(l)}$ is computed as $x^{(l)} = \phi(W^{(l)}x^{(l-1)} + b^{(l)})$, where $W^{(l)}$ and $b^{(l)}$ are the weights and bias, $\phi$ is the nonlinear activation function. $\hat{y}$ is the output of the NN. Assume the dimension of the output of the last layer is $c$. The empirical average loss we want to minimize is: 
\vspace{-2mm}
\begin{equation}
    f(\theta) = \frac{1}{N}\sum_{i=0}^N f_i(\theta) = \frac{1}{N}\sum_{i=1}^N\varepsilon(\hat{y}_i(\theta), y_i)
    \vspace{-1.7mm}
\end{equation}
where $\theta$ denotes all the parameters in the NN, $\theta \in \mathbb{R}^n$ ($n$ is the number of all the parameters) and $\varepsilon(\hat{y}_i(\theta), y_i)$ is a loss function based on the differences between the output $\hat{y}_i$ and the ground truth $y_i$, ($\hat{y}_i, y_i \in \mathbb{R}^c$) for a given dataset $\{(x_1, y_1), (x_2, y_2), ..., (x_N, y_N)\}$. The loss $\varepsilon$ could be Mean Square Error for the regression task and Cross-Entropy for the classification task. 

To derive the generalized Gauss-Newton method, we first examine the Hessian of $f_i(\theta)$ corresponding to a single data point.
\vspace{-3mm}
\begin{equation}
    \frac{\partial^2f_i(\theta)}{\partial \theta} = J_i^TH_iJ_i + \sum_{j=1}^c(\frac{\partial f_i(\theta)}{\partial \hat{y}_i})_j\frac{\partial}{\partial \theta}(J_i)_j
    \vspace{-3mm}
\end{equation}
where $J_i = \frac{\partial \hat{y}_i}{\partial \theta}$ and $H_i = \frac{\partial ^2 f_i(\theta)}{\partial (\hat{y}_i)^2}$. In the Gauss-Newton method, we approximate the Hessian by ignoring the second term and only keeping $J_i^T H_i J_i$ where $J_i \in \mathbb{R}^{c\times n}$ and $H_i \in \mathbb{R}^{c\times c}$. Finally, $\frac{\partial ^2 f_i(\theta)}{\partial \theta^2}$ is approximated by $H = \frac{1}{N}\sum_{i=1}^N J_i^T H_i J_i$ \citep{ren2019efficient}.
When $f_i(\theta) = \varepsilon(\hat{y}_i(\theta), y_i) = \frac{1}{2}\Vert \hat{y}_i(\theta) - y_i\Vert_2^2$, which is the Mean Square Error loss, $H_i = I$, then we have the same formula as stated in the nonlinear least square problem: $H =\frac{1}{N}\sum_{i=1}^N J_i^T J_i.$

When $f_i(\theta) = \varepsilon(\hat{y}_i(\theta), y_i)$ is the Cross Entropy loss, if we assume $\hat{y}_i \in \mathbb{R}^c$ is a vector which represents the probabilities of each class, and $y_i \in \mathbb{R}^c$ is a one-hot encoding vector with 1 for the groundtruth class $k$. Then  $H_i = \frac{\partial ^2 f_i(\theta)}{\partial (\hat{y}_i)^2}$ is a matrix in $\mathbb{R}^{n\times n}$ with only one non-zero entry at the position $(k, k)$, therefore the approximated $H$ can be further simplified to 
\vspace{-3mm}
\begin{equation}
    H = \frac{1}{N}\sum_{i=1}^N \nabla f_i(\theta) \nabla f_i(\theta)^T
    \vspace{-3mm}
\end{equation}
Finally we add the regularization term as we did in the LM  method and get $H = H + 
\lambda I.$

\subsection{Dampening: Maximum Diagonal}
To improve on the Gauss-Newton method, in the LM method we dampen the $J^{T}J$ matrix with a diagonal cut-off. The LM step therefore becomes: $d\theta = -(J^{T}J + \lambda D^{T}D)^{-1}\nabla f(\theta)^{T}.$ Where $D^{T}D$ is a positive-definite matrix representing the relative scaling of the parameters and $\lambda$ is the dampening parameter. We define $D^{T}D$ as the element-wise maximum of the diagonal of all the previous $J^{T}J$ and the current one. We initialize $D^TD = 0.01 I.$ This dampening matrix preserves rescaling invariance while being more robust at avoiding parameter evaporation \citep{transtrum2012improvements}. 
\vspace{-1mm}
\subsection{LM with Adaptive Momentum}
By combining the merits of the Levenberg-Marquardt (LM) method and the Conjugate Gradient (CG) method, one can incorporate an adaptive momentum term into the LM method. And can formulate training a NN as a constrained optimization problem \citep{ampazis2000levenberg}.

We denote the NN parameters $\theta_t$, the updates $d\theta_t$ at time step $t$, and $H_t$ is the (approximated) Hessian. Two update $\theta_t$ and $\theta_{t-1}$ are non-interfering or mutually conjugate w.r.t. $H$ when $d\theta_t^T H d\theta_{t-1} = 0$ Our goal is to maximize 
\vspace{-2mm}
\begin{equation}
    \Phi_t = d\theta_t H_t d\theta_{t-1}
    \vspace{-2mm}
\end{equation}
without compromising the need for a decrease  of the loss function $f(\theta)$ so we add two constraints where $\delta Q_t < 0$.
\vspace{-2mm}
\begin{equation}
    d\theta_t^T H_t d\theta_t = (\delta P)^2 \quad df(\theta_t) = \delta Q_t
    \vspace{-2mm}
\end{equation}
The constrained optimization problem can be solved analytically by introducing Lagrange multipliers $z_1, z_2$
\vspace{-2mm}
\begin{equation}
    \phi_t = \Phi_t + z_1(\delta Q_t - df(\theta_t)) + z_2[(\delta P)^2 - d\theta_t^T H_t d\theta_t]
    \vspace{-2mm}
\end{equation}
To maximize the $\phi_t$ we get the following:
\vspace{-2mm}
\begin{equation}
    d\theta_t = -\frac{z_1}{2z_2}H_t^{-1}\nabla f(\theta_t) + \frac{1}{2z_2}d\theta_{t-1}
    \vspace{-2mm}
\end{equation}
Define the following terms:
\vspace{-2mm}
\begin{equation}
    I_{GG} = \nabla f(\theta_t)^TH_t^{-1}\nabla f(\theta_t) \quad 
    I_{GF} = \nabla f(\theta_t)^Td\theta_t \quad
    I_{FF} = d\theta_{t-1}^T H_t d\theta_{t-1}
    \vspace{-2mm}
\end{equation}
Then $z_1, z_2$ can be evaluated in terms of known quantities
\vspace{-1mm}
\begin{equation}
z_2 = \frac{1}{2}[\frac{I_{GG}(\delta P)^2 - (\delta Q_t)^2}{I_{FF}I_{GG} - I_{GF}^2}]^{-1/2} \quad
z_1 = \frac{-2z_2\delta Q_t + I_{GF}}{I_{GG}}
\vspace{-2mm}
\end{equation}
\subsection{Learning Rate Line Search}

Due to the trust region nature of the LM method, one will have successful iterations as well as failed ones. A successful iteration is defined as an iteration whose updated weight results in a lower loss than the previous iteration. In the case of a failed iteration, we perform a line search over the learning rate (LR) to minimize the loss function. We then update our network using those parameters. The algorithm is produced in Alg. \ref{algo:lr_search} in the appendix. We do not count the line search as extra iterations since we are not recalculating the network's Jacobian or an inverse. 
\subsection{Uphill Steps}
The LM algorithm requires a rule to accept or reject a proposed step. A natural and safe choice is to only accept steps that reduce the loss. But this may not be the most efficient acceptance rule. The Uphill step \citep{transtrum2012improvements} rule accepts steps such that the algorithm is moving in the general direction of the previous iteration. Formally $\beta_i = \text{cos(}d\theta^{new},d\theta^{old}).$ This denotes the cosine angle between the proposed direction $d\theta^{new}$ and the last accepted step $d\theta^{old}.$ In words, we want to accept uphill moves if the angle $\beta_i$ is acute, increasingly as $\beta_i \to 0.$ Specifically the uphill acceptance rule becomes $(1-\beta_i)^b f(\theta^{t+1}) \leq f(\theta^t)$ or more conservatively if $(1-\beta_i)^b f(\theta^{t+1}) \leq \min( f(\theta^i), \cdots, f(\theta^t)).$ As $b$ increases the algorithm will accept uphill steps more freely.
\section{Evaluation and Comparison}

\subsection{Setup}
We use a synthetic noisy Sine dataset and MNIST to evaluate the NN. We use the best hyper-parameters per optimizer. See appendix \ref{experiment} for more details.  
\subsection{Regression Problem}
\begin{figure}[htbp]
    \vspace{-3mm}
    \centering
    \scalebox{0.8}{
    \subcaptionbox{}{\includegraphics[width=0.471\textwidth, angle=0]{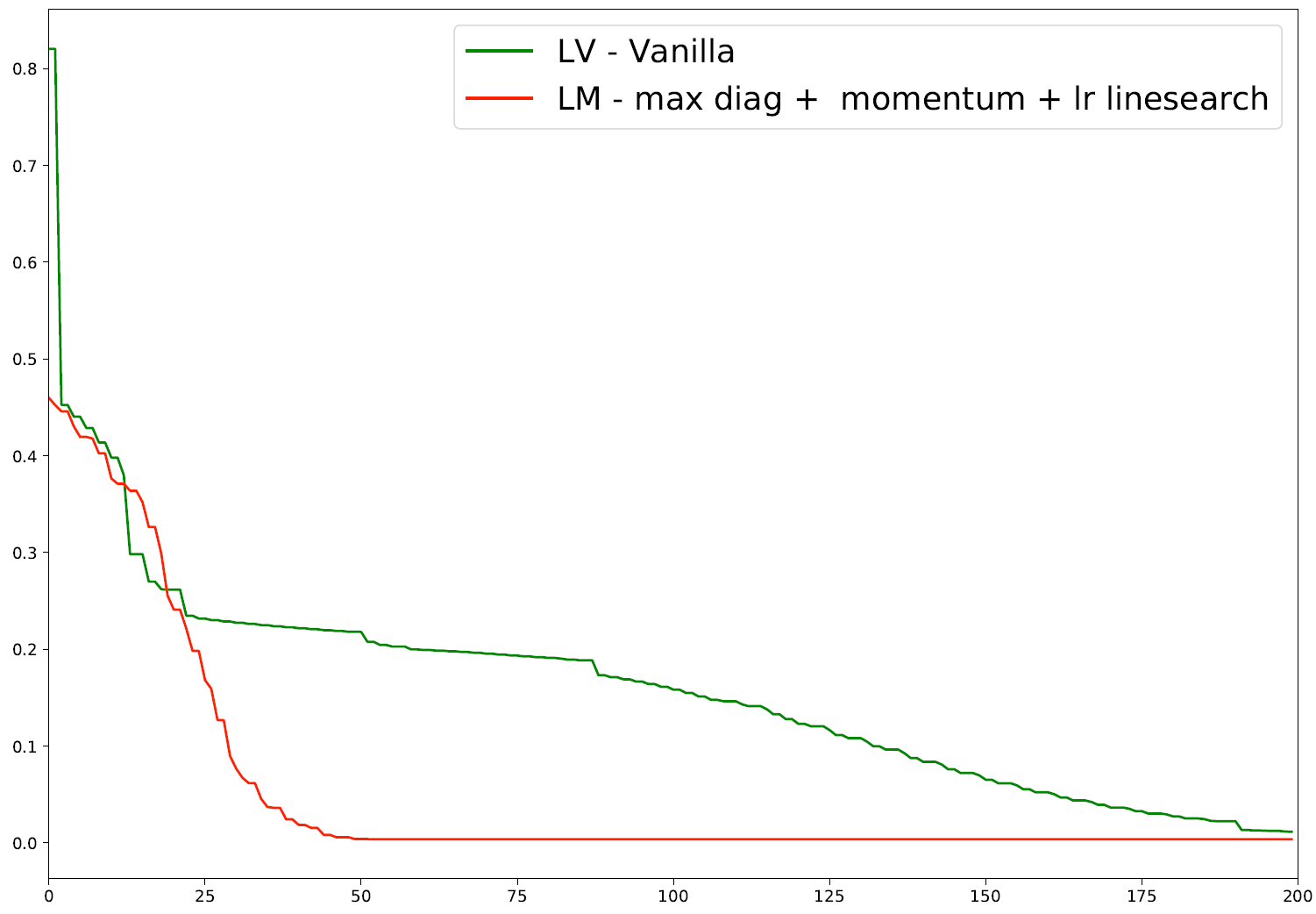}}
    \hspace{10mm}
    \subcaptionbox{}{\includegraphics[width=0.471\textwidth,angle=0]{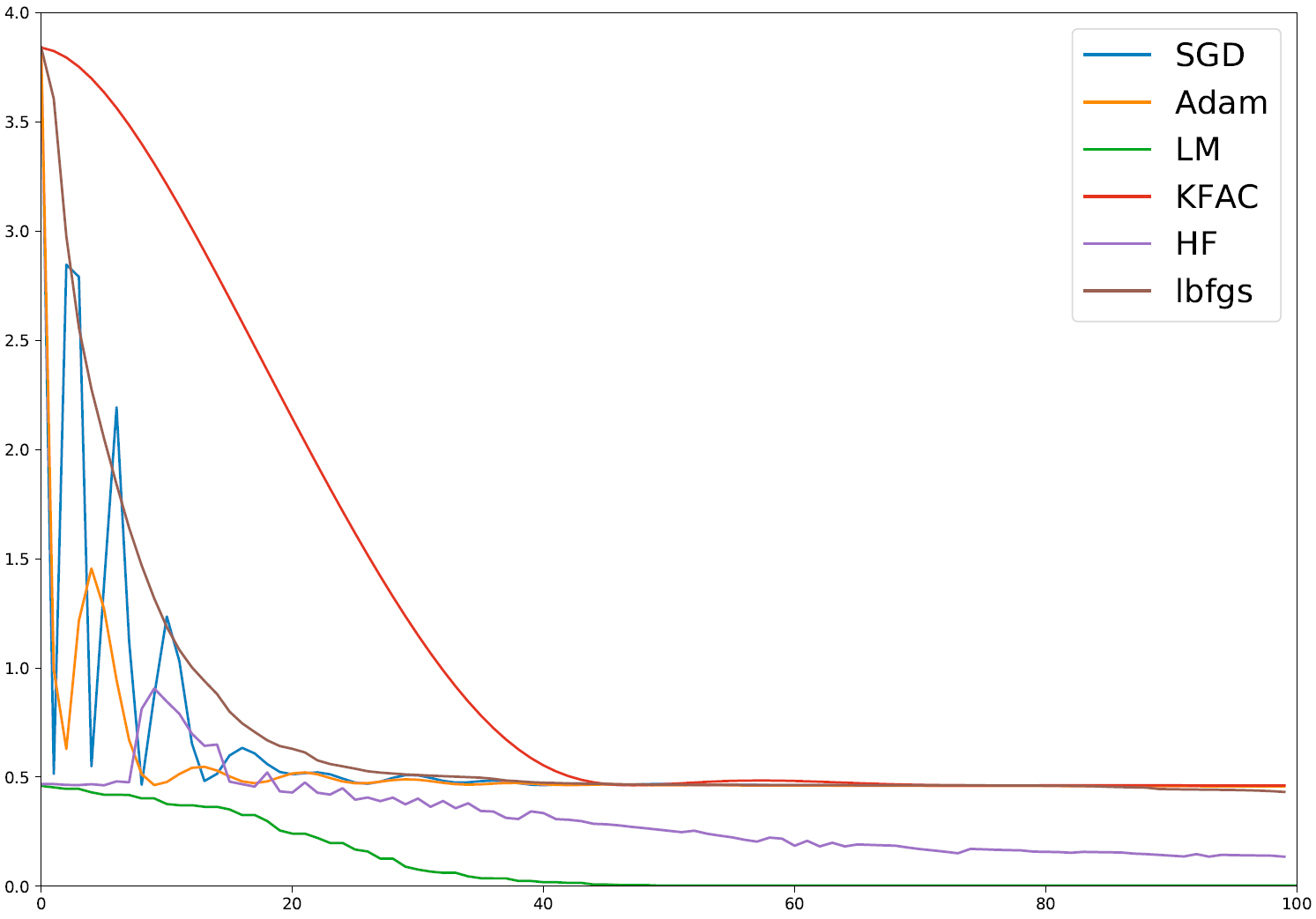}}
    }
    \vspace{-3mm}
    \caption{Comparison of optimizers on Sine regression problem. The X-axis is iterations \&  y-axis is loss. a) vanilla LM vs updated LM. b) Comparing $1^{st}$ and $2^{nd}$ order optimizers vs modified LM}%
    \label{fig:regression_compare}%
    \vspace{-3mm}
\end{figure}

We begin our experiments with a vanilla LM method using the GN approximation of the Hessian matrix. We train a 2-layer MLP with ELU activations to do a regression of a noisy Sine function. We then add adaptive momentum, LR line search, and maximum diagonal dampening in LM, and compared. As we see, in Figure \ref{fig:regression_compare} (a), adding these methods to LM greatly improves the convergence. 

 In Figure \ref{fig:regression_compare} (b) compares modified LM to other first and Quasi/Gauss-Newton methods with tuned hyper-parameters. We see that our modified LM method was able to converge faster and to a lower loss compared to the rest. First order optimizers as well as KFAC converged to a loss of 0.5 whereas LBFGS and HF methods converged to around 0.25. In comparison, LM with max diagonal components of the GN matrix, adaptive momentum, and LR line search converged to 0.001. 

As second-order methods are slower per iteration compared to first-order methods, we train a NN till the loss converges using the LM method and allow other optimizers the same time. In Figure \ref{fig:regression_compare} LM converges to a lower loss representing a better solution to the regression problem (see Figure \ref{fig:regression_compare_sine}).

\subsection{MNIST Classification}
\begin{figure}[ht]
    \vspace{-3mm}
    \centering
    \scalebox{0.7}{
    \vspace{-2mm}
    \subcaptionbox{Training Loss \label{subfige:forget}}{    \includegraphics[width=.475\textwidth,trim=0 -2mm 0 0]{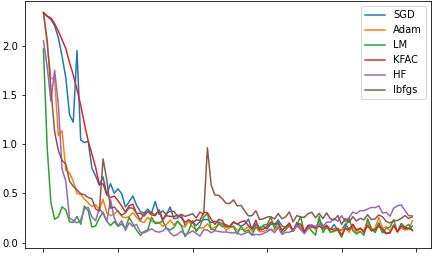}}
    \hfill 
    \subcaptionbox{Training Accuracy \label{subfige:certain}}{     \includegraphics[width=.475\textwidth,trim=0 0 0 0]{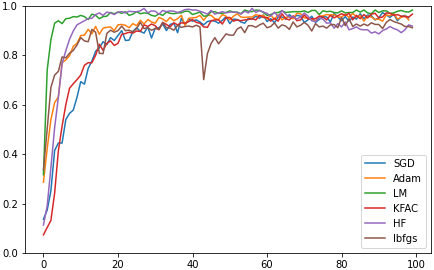}}    
    }
    \medskip
    \vspace{-2mm}
    
    \scalebox{0.7}{
    \subcaptionbox{Testing Loss \label{subfige:selected}}{	\includegraphics[width=.475\textwidth,trim=0 -2mm 0 0]{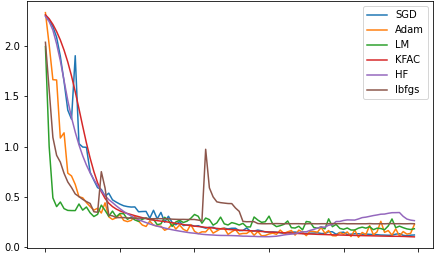}}

    \hfill 
    \subcaptionbox{Testing Accuracy \label{subfige:not_selected}}{    \includegraphics[width=.475\textwidth,trim=0 0 0 0]{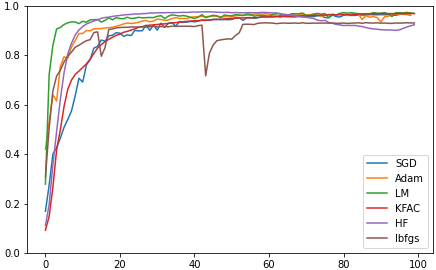}
    }
    }
    \vspace{-3mm}
    \caption{CNN on Full MNIST (100 Epochs)
    }
    \label{fig:compare_mnist}
    \vspace{-4mm}
\end{figure}

Next, we train a Convolutional Neural Network on the MNIST dataset. In Figure \ref{fig:compare_mnist} the LM method optimizes the NN in fewer epochs, converging to a lower loss with higher test and train accuracy compared to the rest of the optimizers. In our experiments, HF optimization tends to be unstable. We adjust the learning rate as well as increased the number of CG iterations for the HF method but we often found instabilities with both HF and LBFGS. We found LBFGS would often get stuck at an accuracy of 1\%; we omit these runs from the results.

\vspace{-1mm}
Table \ref{table:timing} is the overall timing and results of the full MNIST experiment on a CNN. We ran each optimizer and timed them till they reached an accuracy of 90\%. We see that the LM method reaches the best loss followed by Adam. Yet the time it took to run the first-order optimizers was less than other methods. LM converges to a lower loss and better accuracy than the other optimizers, even if it is slower. As we also observed from the noisy Sine regression before, the LM method can learn well from a small amount of data, as shown by the few numbers of iterations it took to reach the 90\% accuracy mark. 
\begin{table}[]
\centering
\vspace{-5mm}
\scalebox{0.65}{
\begin{tabular}{@{}lllllll@{}}
\toprule
\textbf{}                                  & \textbf{LM} & \textbf{KFAC}         & \textbf{HF}             & \textbf{LBFGS} & \textbf{Adam} & \textbf{SGD} \\ \midrule
\multicolumn{1}{l|}{Time}                  & 0.9         & \multicolumn{1}{c}{3} & \multicolumn{1}{c}{4.7} & 5.6            & 0.1           & 0.1          \\
\multicolumn{1}{l|}{Epochs to reach 90\%} & 4   & 26  & 11 & 17  & 13  & 26  \\
\multicolumn{1}{l|}{Converged}            & Yes & Yes & No          & Yes & Yes & Yes \\
\multicolumn{1}{l|}{Accuracy (100 Epochs)} & 97.5\%      & 96\%               & 94\%                & 91\%           & 96.5\%         & 96\%          \\ \bottomrule
\end{tabular}
}
\vspace{-2.5mm}
\caption{Comparison of Timing, variance in appendix}
\label{table:timing}
\vspace{-5mm}
\end{table}
Figure \ref{fig:ablation} shows the ablation study on LM and demonstrates the effectiveness of learning rate line search, adaptive momentum, as well as uphill techniques. In Figure \ref{fig:ablation}, we see that these adjustments significantly improve the stability of the loss and improve accuracy.\
\vspace{-2mm}
\section{Conclusion and Future Work}
We implement the Levenberg-Marquardt method for neural networks and observe favorable convergence performance on both regression and classification tasks. Furthermore, learning rate line search, adaptive momentum, and uphill steps are all useful techniques to improve the performance of the LM method. With this in mind, we found this method to not be scalable due to memory and computational constraints. We believe that some of the methods explored in this paper can be applied to very recently proposed, cheap curvature-aware optimizers such as PSGD \citep{xilinpsgd,xilinonline,xilinpsgdlie,li2022black}.

\clearpage

\bibliography{main}
\appendix

\newpage
\section{Theoretical Background} \label{appendix}
\subsection{Newton Method}
Newton's method is a second-order method which enjoys fast convergence, robustness and affine invariance. In a single step, instead of just advancing towards the minimum, it steps towards the global minimum under the assumption that the function is actually quadratic and its second order expansion is a good approximation. This allows the optimizer to make big steps in low-curvature scenarios and small steps in high-curvature scenarios. The Newton step $v_{\text{nt}}$ is calculated by minimizing the second-order approximation in \eqref{second_approx}. However, computing and storing the Hessian $H = \nabla ^2f(x)$ and its inverse is expensive in computation (time complexity $O(n^3)$) and memory (space complexity $O(n^2)$). Several variations are proposed to approximate the Hessian $H$ more efficiently.
\begin{equation}
    \hat{f}(x+v) = f(x) + \nabla f(x)^Tv + \frac{1}{2}v^T\nabla^2 f(x)v 
    \label{second_approx}
\end{equation}

\begin{equation}
    v_{\text{nt}} = -\nabla^2 f(x)^{-1} \nabla f(x)
\end{equation}

\subsection{Quasi-Newton Method}

Quasi-Newton method uses inverse Hessian estimation to compute each update iteration. The most popular one is the Broyden–Fletcher–Goldfarb–Shanno (BFGS) update. 
\begin{equation}
    H_{k+1}^{-1} = (I-\frac{sy^T}{y^Ts}) H_k^{-1} (I-\frac{ys^T}{y^Ts}) + \frac{ss^T}{y^Ts}
\end{equation}

\begin{equation}
    s = x_{k+1} - x_k \quad y = \nabla f(x_{k+1}) -\nabla f(x_k)
\end{equation}

where $H_k^{-1}$ is the approximation of inverse Hessian at iteration $k$. Then to further tackle the memory issue, limited-memory BFGS (L-BFGS) was proposed to avoid storing $H_k$ and $H_k^{-1}$ by evaluating $H_k$ recursively for $m$ iterations ($j = k-1, ..., k-m$).
\begin{equation}
    H_{j+1}^{-1} = (I-\frac{s_jy_j^T}{y_j^Ts_j}) H_j^{-1} (I-\frac{y_js_j^T}{y_j^Ts_j}) + \frac{s_js_j^T}{y_j^Ts_j}
\end{equation}

\begin{equation}
    s_j = x_{j+1} - x_j \quad y_j = \nabla f(x_{j+1}) -\nabla f(x_j)
\end{equation}

\subsection{Gauss-Newton Method}

Gauss-Newton method was originally proposed to solve the nonlinear least squares problems, where $r: \mathbb{R}^n \rightarrow \mathbb{R}^m$ denotes the residual $r(x) = [r_1(x), ..., r_m(x)]^T$ of a $n$-vector $x$.
\begin{equation}
    \min \quad g(x)= \Vert r(x) \Vert_2^2 = \sum_{i=0}^{m}r_i(x)^2
\end{equation}

To optimize the function above, we can take a Newton step, which will do the following update
\begin{equation}
    v_{\text{nt}} = -\nabla^2g(x)^{-1}\nabla g(x) = -(J^T J + \sum_{i=1}^{m}r_i(x)\nabla^2r_i(x))^{-1}J^Tr
\end{equation}

where $J = \frac{\partial r(x)}{\partial x}$ is the Jacobian matrix of $r(x)$. However in the Gauss-Newton step, we discard the second term in the inverse and simply do
\begin{equation}
    v_{\text{gn}} = -(J^TJ)^{-1}J^Tr
\end{equation}

So the Gauss-Newton step doesn't require second derivatives and always provides a descent direction when $J$ has full column rank; because $H=2J^TJ$ is positive definite.

Levenberg-Marquardt algorithm is a variant of Gauss-Newton method which addresses the indefiniteness of the approximated Hessian by adding a regularization term $\lambda$
\begin{equation}
    v_{\text{gn}} = -(J^TJ + \lambda I)^{-1}J^Tr
\end{equation}

Here $\lambda$ essentially defines a trust region where the quadratic approximation is a good fit. It blends the linear and quadratic approximations. When $\lambda = 0$, the update rule is completely based on the second-order approximation, when $\lambda$ is very large, the rule is completely based on the first-order approximation.

\subsection{Generalized Gauss-Newton Method}

Consider a neural network (NN) with $L+1$ layers, at the $l$-th layer, given the outputs from the preceding layer $x^{(l-1)}$ as input, $x^{(l)}$ is computed as $x^{(l)} = \phi(W^{(l)}x^{(l-1)} + b^{(l)})$, where $W^{(l)}$ and $b^{(l)}$ are the weight and bias, $\phi$ is the nonlinear activation function. $\hat{y}$ is the output of the NN. Assume the dimension of output of the last layer is $c$. The empirical average loss we want to minimize is: 
\begin{equation}
    f(\theta) = \frac{1}{N}\sum_{i=0}^N f_i(\theta) = \frac{1}{N}\sum_{i=1}^N\varepsilon(\hat{y}_i(\theta), y_i)
\end{equation}

where $\theta$ denotes all the parameters in the NN, $\theta \in \mathbb{R}^n$ ($n$ is the number of all the parameters) and $\varepsilon(\hat{y}_i(\theta), y_i)$ is a loss function based on the differences between the output $\hat{y}_i$ and the groudtruth $y_i$, ($\hat{y}_i, y_i \in \mathbb{R}^c$) for a given dataset $\{(x_1, y_1), (x_2, y_2), ..., (x_N, y_N)\}$. The loss $\varepsilon$ could be Mean Square Error for the regression task and Cross Entropy for the classification task. 

To derive the generalized Gauss-Newton method, we first examine the Hessian of $f_i(\theta)$ corresponding to a single data point.
\begin{equation}
    \frac{\partial^2f_i(\theta)}{\partial \theta} = J_i^TH_iJ_i + \sum_{j=1}^c(\frac{\partial f_i(\theta)}{\partial \hat{y}_i})_j\frac{\partial}{\partial \theta}(J_i)_j
\end{equation}

where $J_i = \frac{\partial \hat{y}_i}{\partial \theta}$ and $H_i = \frac{\partial ^2 f_i(\theta)}{\partial (\hat{y}_i)^2}$. In the Gauss-Newton method, we approximate the Hessian by ignoring the second term and only keeping $J_i^T H_i J_i$ where $J_i \in \mathbb{R}^{c\times n}$ and $H_i \in \mathbb{R}^{c\times c}$. Finally, we approximate $\frac{\partial ^2 f_i(\theta)}{\partial \theta^2}$ is approximated by\citep{ren2019efficient}
\begin{equation}
    H = \frac{1}{N}\sum_{i=1}^N J_i^T H_i J_i
\end{equation}

When $f_i(\theta) = \varepsilon(\hat{y}_i(\theta), y_i) = \frac{1}{2}\Vert \hat{y}_i(\theta) - y_i\Vert_2^2$, which is the Mean Square Error loss, $H_i = I$, then we have the same formula as stated in the nonlinear least square problem.
\begin{equation}
    H =\frac{1}{N}\sum_{i=1}^N J_i^T J_i
\end{equation}

When $f_i(\theta) = \varepsilon(\hat{y}_i(\theta), y_i)$ is the Cross Entropy loss, if we assume $\hat{y}_i \in \mathbb{R}^c$ is a vector which represents the probabilities of each class, and $y_i \in \mathbb{R}^c$ is a one-hot encoding vector with 1 for the groundtruth class $k$. Then  $H_i = \frac{\partial ^2 f_i(\theta)}{\partial (\hat{y}_i)^2}$ is a matrix in $\mathbb{R}^{n\times n}$ with only one non-zero entry at the position $(k, k)$, therefore the approximated $H$ can be further simplified to 
\begin{equation}
    H = \frac{1}{N}\sum_{i=1}^N \nabla f_i(\theta) \nabla f_i(\theta)^T
\end{equation}

Finally we add the regularization term as we did in the LM  method and get $H = H + 
\lambda I$

\subsection{Hessian-Free}

Hessian-free optimization uses the insights from Newton's method but comes up with a better way to minimize the quadratic function. Just as in Newton's method, given a function $f$, we find the second-order Taylor expansion:
\begin{equation}
f(x+v) \approx f(x)+\nabla f(x)^{T} v + v^{T} H(f) v    
\end{equation}

We then find the minimum of this approximation (the best $\Delta x$) using conjugate gradient, take an iterative step to $x = x+v$, and continue this iteration until we have convergence.

\begin{algorithm}[H]
\caption{Hessian-Free}
\label{algo:hf}
Let $f$ be any function $f: \mathbb{R}^{n} \rightarrow \mathbb{R}$ which we wish to minimize.\\
    \begin{enumerate}
        \item Initialize:\\ 
        ~Let $i=0$ and $x_{i}=x_{0}$ be some initial guess.\\
        \item  Set up quadratic expansion:\\ 
        ~At the current $x_{n}$, compute the gradient $\nabla f\left(x_{n}\right)$ and Hessian $H(f)\left(x_{n}\right),$ and consider the following Taylor expansion of $f:$\\
        \[
f(x+ v) \approx f(x)+\nabla f(x)^{T} v+v^{T} H(f) v
\]
        \item Conjugate gradient:\\ 
        ~Compute $x_{n+1}$ using the Conjugate Gradient algorithm for quadratic functions on the current Taylor expansion.\\
        \item  Iterate:\\ 
        ~Repeat steps 2 and 3 until $x_{n}$ has converged.
    \end{enumerate}
\end{algorithm}

Fortunately in the Hessian free algorithm, it is possible to circumvent the full Hessian expansion in step 2, and instead only requires us to calculate the much cheaper Hessian vector product $Hv$.\\

To see this, consider the first component of $Hv$. The $i-$th row of the Hessian contains partial derivatives of the form $\frac{\partial^2}{\partial x_i x_j} f$ where $j$ is the column index. As per normal matrix-vector multiplication, the $i-$th row of $Hv$ is the dot product of $v$ and $i$th row of $H$. The $i-$th row of $H$ can be expressed as the gradient of the derivative of $f$ with respect to $x_i$, yielding \citep{gibanskyhf}:
\[(Hv)_i = \sum_{j=1}^N \frac{\partial^2f}{\partial x_i x_j}(x) \cdot v_j = \nabla \frac{\partial f}{\partial x_i}(x) \cdot v
\]

One can utilize auto-differentiation packages to calculate the exact Hessian vector product as detailed by \citep{pearlmutter1994fast} The operator is defined\\
\[
\left.\mathcal{R}_{\mathbf{v}}\{f(\mathbf{x})\} \equiv \frac{\partial}{\partial t} f(\mathbf{x}+t \mathbf{v})\right|_{t=0}
\]
so $\left.\mathbf{H} \mathbf{v}=\mathcal{R}_{\mathbf{v}}\left\{\nabla_{\mathbf{x}}(\mathbf{x})\right\} . \text { (To avoid clutter we will usually write } \mathcal{R}\{\cdot\} \text { instead of } \mathcal{R}_{\mathbf{v}}\{\cdot\} \right)$ We can then
take all the equations of a procedure that calculates a gradient, $e . g$. the backpropagation procedure, and we can apply the $\mathcal{R}_{v}\{\cdot\}$ operator to each equation. Because $\mathcal{R}\{\cdot\}$ is a differential operator, it obeys the usual rules for differential operators, such as:
\[
\begin{aligned}
\mathcal{R}\{c f(\mathbf{x})\} &=c \mathcal{R}\{f(\mathbf{x})\} \\
\mathcal{R}\{f(\mathbf{x})+g(\mathbf{x})\} &=\mathcal{R}\{f(\mathbf{x})\}+\mathcal{R}\{g(\mathbf{x})\} \\
\mathcal{R}\{f(\mathbf{x}) g(\mathbf{x})\} &=\mathcal{R}\{f(\mathbf{x})\} g(\mathbf{x})+f(\mathbf{x}) \mathcal{R}\{g(\mathbf{x})\} \\
\mathcal{R}\{f(g(\mathbf{x}))\} &=f^{\prime}(g(\mathbf{x})) \mathcal{R}\{g(\mathbf{x})\} \\
\mathcal{R}\left\{\frac{d f(\mathbf{x})}{d t}\right\} &=\frac{d \mathcal{R}\{f(\mathbf{x})\}}{d t}
\end{aligned}
\]
Also note that
\[
\mathcal{R}\{\mathbf{x}\}=\mathbf{v}
\]
These rules allow one to use existing auto-differentiation packages to calculate the vector $\mathcal{R}\{\nabla_{\mathbf{x}}\}$ which is precisely the vector  which we desire $Hv$.

In practice, HF sees fewer applications than SGD because its updates are much more expensive to compute, as they involve running linear conjugate gradient (CG) for potentially hundreds of iterations. Each of these iterations requires a matrix-vector product with the curvature matrix (which are as expensive to compute as the stochastic gradient on the current mini-batch). Next, HF’s estimate of the curvature matrix must remain fixed while CG iterates. This means that the method is able to go through much less data than SGD can in a comparable amount of time, making it less well suited to stochastic optimizations.

\subsection{KFAC}
To get the full understand behind the KFAC method one needs to dive into ample theory covered in the paper. Instead of reproducing this, I will instead provide a sketch for the KFAC method in the Newton Step, we well as talk about some differences between KFAC and other methods. The following notes on KFAC are reproduced from \cite{yaroslavvbkfac} for completion. 
 
\subsubsection{Linear Network}
Given a matrix parameter $W$ and a goal of modeling Y as follows 
\begin{equation}
Y = B^{T}WA    
\end{equation}

We define a prediction error as 
\begin{equation}
    r = \hat{Y} - B^{T}WA.
\end{equation}

We minimize the prediction error by minimizing over $W$ in the trace loss
\begin{equation}
    f = \frac{1}{2}tr(r^{T}r)\\
\end{equation}

To minimize we take the first derivative:
\begin{equation}
    df = tr(r^{T}dr) = tr(r^{T}B^{T}dWA)
\end{equation}

By using the trace properties we find the derivative of $W$
\begin{equation}
    G = dW =  -BrA^{T}
\end{equation}

Taking the derivative of $G$ we have:
\begin{equation}
    dG = -BB^{T}dWAA^{T}
\end{equation}

We vectorize both sides and apply the Kronecker-Vector transform rule:
\begin{equation}
    vec(dG) = (AA^{T} \otimes BB^{T})vec(dW)
\end{equation}

We can then by inspection see 
\begin{equation}
    H = (AA^{T} \otimes BB^{T})
\end{equation}

Since we have extracted the Hessian from an equation that was vectorized we write our update rule as:
\begin{equation}
    vec(W) - H^{-1}vec(G)
\end{equation}

To calculate this inverse we note that the inverse distributes over Kronecker product and the fact that vec distributes over Kronecker product we have
\begin{equation}
    vec(W - (BB^{T})^{-1}G(AA^{T})^{-1}).
\end{equation}

Which gives us the Newton update step for the unvectorized form of the problem. \\
\subsubsection{Piecewise Linear Network}
For a piecewise linear neural network, the loss for each example is locally linear, and we can write our total loss as sum of per-example losses\\
\begin{equation}
    f = \sum_i f_i
\end{equation}

where each example $i$ is associated with its own versions of $A_i, B_i$ and $r_i.$ And so we have the per-example derivative of the loss:
\begin{equation}
    df_i = -tr(r_{i}^{T}B_{i}^{T}dW_{i}A_{i})
\end{equation}

Where the toal Hessian is the sum of $H_i$ which we can approximate using equasion 1 in the KFAC paper
\begin{equation}
    H = \sum_i (A_iA_i^{T} \otimes B_iB_i^{T}) \approx \sum_i (A_iA_i^{T}) \otimes \sum_i (B_iB_i^{T})
\end{equation}

Since each example corresponds to a vector we can stack these vectors into a matrix resulting in $H \approx AA^{T} \otimes BB^{T} $. Again, our Hessian inverse can be written as
\begin{equation}
    H^{-1} \approx  (AA^{T})^{-1} \otimes (BB^{T})^{-1}
\end{equation}

\subsubsection{Intuition and Distinguishing Factors of KFAC}

The core of the KFAC method is to use an approximation of the Fisher Information matrix which assumes each of the layers are independent. The independence can be seen by the authors formulation in which all the weights of a each layer of a NN are broken down into groups by rows and columns in the Fisher matrix. This allows for the Fisher matrix to be factored into the multiplication of much smaller matrix multiplications via Kronecker factorization. Such a factorization allows for very efficient inversion of the Fisher matrix. Online estimation of the matrix is possible using arbitrarily large subsets of the training data (without increasing the cost of inversion).This assumption, although a strong, allows for the math to be factored in a computationally efficient way.  While the assumption that the layers of a NN are independent is a strong one, it allows for the math to be factored in a computationally efficient way.

Another distinguishing factor that sets the KFAC method apart from the rest of the optimization methods looked at thus far, is that the KFAC method can estimate the curvature matrix from a lot of data by using an online exponentially-decayed average, as opposed to relatively small-sized fixed mini-batches used by the other stochastic methods. This allows for curvature estimates to depend on much more data than can be reasonably processed in a single mini-batch. 

Notably, for methods like LM which deals with an approximate Hessian directly or HF which deal with the exact Fisher indirectly via matrix-vector products, such a scheme would be impossible to implement efficiently, as the exact Fisher matrix (or Generalized-Gauss-Newton) seemingly cannot be summarized using a compact data structure whose size is independent of the amount of data used to estimate it. Indeed, it seems that the only representation of the exact Fisher which would be independent of the amount of data used to estimate it would be an explicit $n \times n$ matrix (which is far too big to be practical). Because of this, HF and related methods must base their curvature estimates only on subsets of data that can be reasonably processed all at once, which limits their effectiveness in the stochastic optimization regime \citep{martens2015optimizing}.

\subsection{Learning Rate Line Search}
    \begin{algorithm}[H]
        \caption{Learning Rate Line Search }
        \label{algo:lr_search}
        Initialize lrs = (1e-6,9) with step size of 0.125, empty list loss\_list = []  \;\\
        \For{lr $\in$ lrs }{
            Update network: $\theta^{(t+1)} = \theta^{(t)} + lr* v_{gn}$\\
            Evaluate network: $loss_{lr} = f(\theta^{(t+1)})$\\
            Add current loss to list: loss\_list.append($loss_{lr}$)\\
            Reset network to $\theta^{t}$
        }
        lr\_best = $\argmin(\text{loss\_list})$\\
        Update network with lr of least loss: $\theta^{(t+1)} = \theta^{(t)} + \text{lr\_best}* v_{gn}$
        
    \end{algorithm}

\section{Experimental Setup} \label{experiment}
For the regression task, we fit the Sine function with a simple two-layer fully-connected MLP, ELU as the non-linear activation function, and mean square error as the loss function. 
For the classification task, we use the MNIST datasets, with $28\times28$ resolution, 60000 training samples and 10000 testing samples. We train a convolutional NN with two convolution layers, and Randomized Leaky ReLu as the non-linear activation function, two max-pooling layers and one fully-connected layers. We compare LM method with popular first-order methods (SGD, Adam) and second-order methods (L-BFGS, Hessian-Free, KFAC). We use the best hyper-parameters per optimizer. See appendix for more details.

We set the hyper-parameters to be the best for each optimizer.  For SGD we set the learning rate (lr) to be 0.05, with momentum=0.9 and weight decay=5x$10^{-4}$, for Adam, we set the lr to be 0.01, for L-BFGS, we set the lr to  0.005, for Hessian Free, we set the lr to be 0.1, for KFAC, we set the lr to 0.05. For LM, we set the initial lr to be 1 and initial $\lambda$ to  1 as well. For the full sized MNIST problem we use KFAC with a lr of 1, Adam with a lr of 0.05, SGD with a lr of 0.1. Keeping the rest of the learning rates the same. 

\section{Further Experiments}

Wee see that LM can learn the regression problem with a single iteration compared to the results of Adam and KFAC.

\begin{figure}[htbp]
    \centering
    \subcaptionbox{}{\includegraphics[width=0.3\textwidth, angle=0]{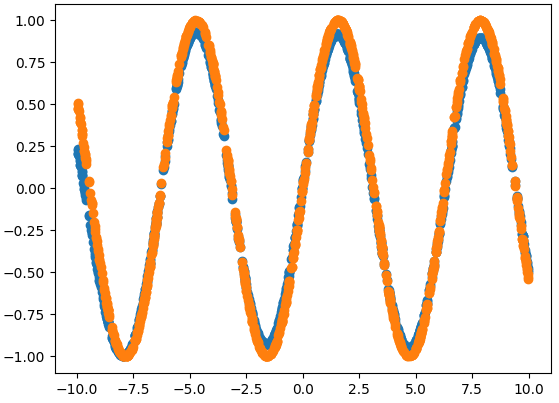}}
    \subcaptionbox{}{\includegraphics[width=0.3\textwidth, angle=0]{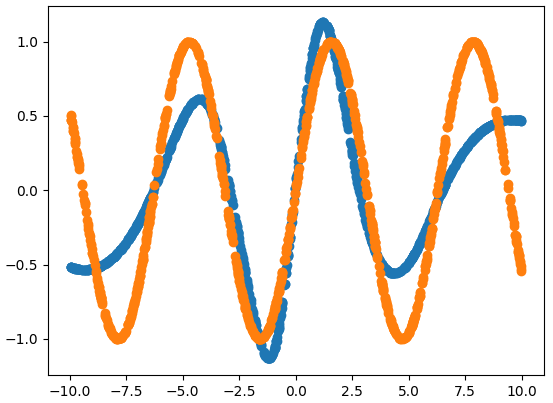}}
    \subcaptionbox{}{\includegraphics[width=0.3\textwidth, angle=0]{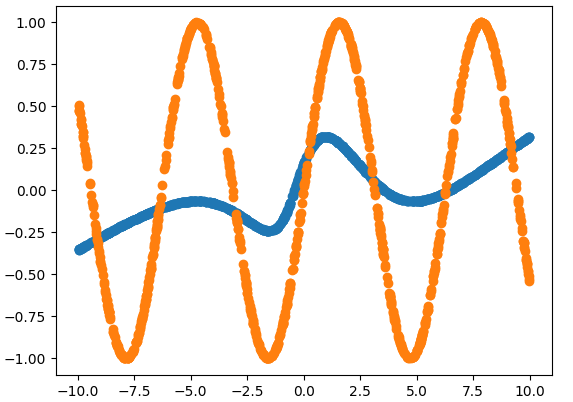}}
    \caption{Comparison of optimizers on simple regression problem a) LM b) KFAC c) Adam}%
    \label{fig:regression_compare_sine}%
\end{figure}

To then see how the optimizers would do  with a limited dataset, we swtiched the train and  the test set and reran the algorithms. We again found that the LM method outperformed the rest, both in the few number of iterations it took to reach a high accuracy and overall accuracy at the end of an epoch.
\begin{figure}[ht]
    \centering
    \includegraphics[width=\textwidth]{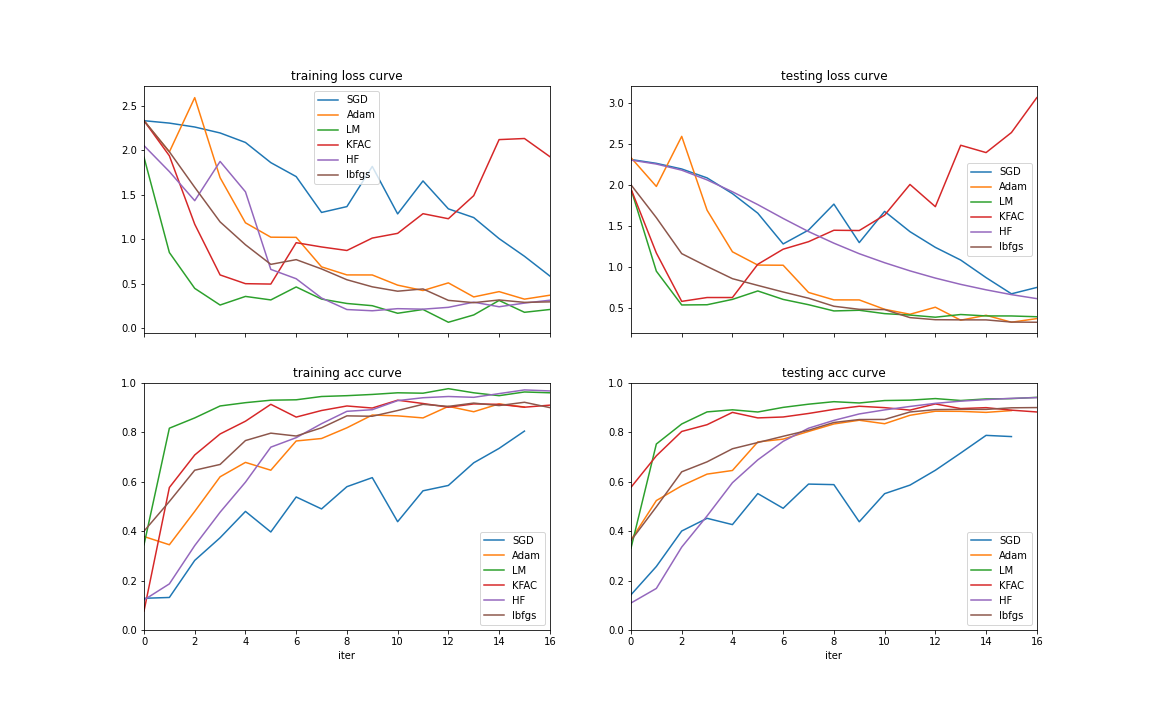}
    \caption{CNN on Full MNIST 1 Epoch Train and Test set switched}
    \label{fig:full_mnist_train}
\end{figure}

\vspace{-5mm}
\begin{figure}[ht]
    \centering
    \scalebox{0.75}{
    \subcaptionbox{Training Loss \label{subfige:forget}}{\includegraphics[width=.475\textwidth,trim=0 0 0 0]{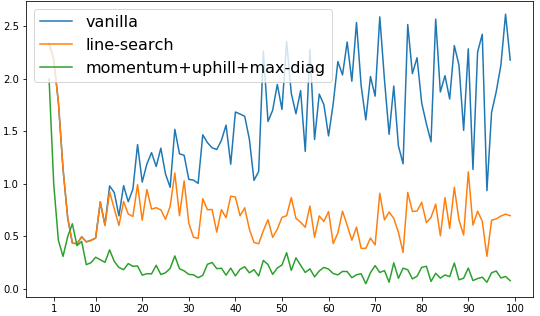}}
    
    \hfill 
    \subcaptionbox{Training Accuracy \label{subfige:certain}}{    \includegraphics[width=.475\textwidth,trim=0 0 0 0]{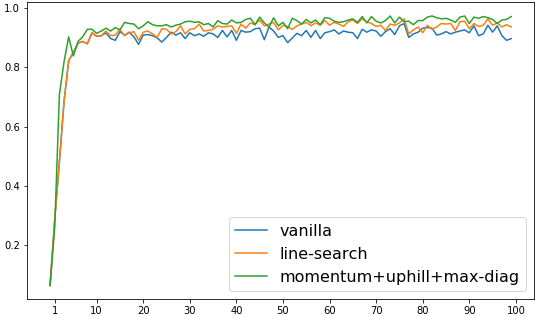}}
    \vspace{-10mm}
    }
    \medskip
    \vspace{-5mm}
    \scalebox{0.75}{
    \subcaptionbox{[Testing Loss \label{subfige:selected}}{\includegraphics[width=.475\textwidth,trim=0 0 0 0]{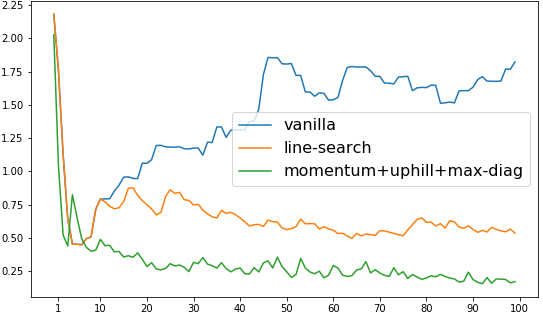}}
    \hfill 
    \subcaptionbox{Testing Accuracy \label{subfige:not_selected}}{\includegraphics[width=.475\textwidth,trim=0 0 0 0]{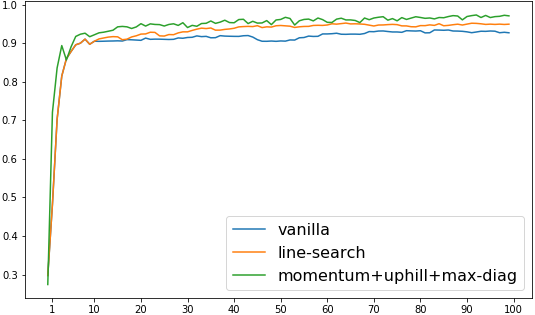}}
    }
    \vspace{-5mm}
    \caption{Ablation Study: CNN on Full MNIST (100 Epochs)
    }
    \label{fig:ablation}
    \vspace{-5mm}
\end{figure}

\section{Future Work}
Very recently \cite{adacore} has shown that one can efficiently learn from large datasets by selecting subsets based on curvature information to train on. In the experiments they see that using curvature informed optimizers can significantly boost performance. We believe by applying our LM method or PSGD to optimize networks with these AdaCore subsets, we can achieve better generalization compared to first order methods. 

Furthermore very recently \cite{causal} sees the use of second order methods to greatly benefit optimization of causal models. We believe there is potential to combine ideas from this paper with more efficient methods like PSGD and apply them to causal models and many other machine learning frameworks. 

\end{document}